\documentclass[]{article}

\usepackage[margin = 2.5cm]{geometry}
\usepackage{parskip}

\usepackage{algorithm}
\usepackage{algpseudocode}

\usepackage{bm}
\usepackage{mathtools}
\usepackage{amsfonts}
\usepackage{amsmath}
\usepackage{amsthm}

\usepackage{float}
\usepackage{graphicx}
\usepackage{xcolor}
\usepackage{tikz}

\usepackage[compress]{natbib}

\renewcommand{\P}{\mathbb{P}}

\newcommand{\E}{\mathbb{E}}

\newcommand{\defeq}{\vcentcolon =}

\newcommand{\R}{\mathbb{R}}

\newcommand{\eg}{\emph{e.g.},~}
\newcommand{\ie}{\emph{i.e.},~}

\renewcommand{\eqref}[1]{Eq.~(\ref{#1})}
\newcommand{\figref}[1]{Fig.~\ref{#1}}
\newcommand{\secref}[1]{Sec.~\ref{#1}}
\newcommand{\appref}[1]{App.~\ref{#1}}
\renewcommand{\algref}[1]{Alg.~\ref{#1}}

\definecolor{dred}{rgb}{0.6,0,0}
\definecolor{dblack}{rgb}{0,0,0}
\definecolor{dgreen}{rgb}{0.1,0.6,0.1}
\usepackage[compress]{natbib}
\usepackage[colorlinks=true,linkcolor=dblack,citecolor=dred, urlcolor=dgreen]{hyperref}
\usepackage[nameinlink,capitalise]{cleveref}
\usepackage{authblk}

\renewcommand{\L}{\mathcal{L}}
\renewcommand{\Pr}{\mathrm{Profit}}
\usepackage{bbm}
\newcommand{\1}{\mathbbm{1}}

\title{Reinforcement Learning applied to Insurance Portfolio Pursuit}

\author[1,2]{Edward James Young\footnote{Corresponding author - \texttt{ey245@cam.ac.uk, edwardjamesyoung3@gmail.com}}}
\author[1]{Alistair Rogers}
\author[1]{Elliott Tong}
\author[1]{James Jordon\footnote{Senior author - \texttt{james.jordon@accenture.com}}}
\affil[1]{Accenture (UK) Limited}
\affil[2]{University of Cambridge, Computational and Biological Learning (CBL) group.}

\date{}

\begin{document}

\maketitle

\begin{abstract}
	When faced with a new customer, many factors contribute to an insurance firm's decision of what offer to make to that customer. In addition to the expected cost of providing the insurance, the firm must consider the other offers likely to be made to the customer, and how sensitive the customer is to differences in price. Moreover, firms often target a specific portfolio of customers that could depend on, \eg age, location, and occupation. Given such a target portfolio, firms may choose to modulate an individual customer's offer based on whether the firm desires the customer within their portfolio. We term the problem of modulating offers to achieve a desired target portfolio the \emph{portfolio pursuit problem}. Having formulated the portfolio pursuit problem as a sequential decision making problem, we devise a novel reinforcement learning algorithm for its solution. We test our method on a complex synthetic market environment, and demonstrate that it outperforms a baseline method which mimics current industry approaches to portfolio pursuit. 
\end{abstract}

\section{Introduction}

\subsection*{Background}

Competition among the commercial insurance market has been intensified by the emergence of price comparison websites (PCW) as a core means by which customers and insurance firms interact. When a customer's details are entered into the website, many firms simultaneously decide what offers to make to the customer on the basis of those details. Since the customer can directly compare competing offers, firms must carefully calibrate their offers to maximise a trade-off between potential profit and the probability of acceptance. While both of these aims are local to each individual customer, firms may also wish to modulate the offers they make in order to pursue longer-term goals, realised over many customers. For example, the firm may wish to maintain a certain frequency with which new offers are accepted, to ensure they have the reserves necessary to pay out claims and avoid liquidity issues \citep{krasheninnikova_reinforcement_2019}. Alternatively, a firm may wish to break into a new segment of the market, and be content with making lower-than-optimal offers in order to attract a particular customer base. Goals such as these define a higher-level strategy that is crucial to the long-term success of the firm. Decisions regarding content of the long-term goals - the \emph{What?} of the strategy - are typically made at a managerial level, and involve complex domain-expert knowledge and understanding of current market conditions. At present, the implementation of the pursuit of the goals - the \emph{How?} of the strategy - also makes use of domain-expect knowledge. In this paper, we present a data-driven machine learning methodology for implementing higher-level strategies in the context of consumer insurance markets. 

The problem of what offer to make to a customer on a PCW can be divided into two sub-problems: the \emph{cost-estimation problem} and the \emph{bidding problem}. The \emph{cost-estimation problem} can be summarised as the problem of estimating the total cost of insuring a given customer from the features available through the PCW. This is a well-studied supervised-learning problem \citep{de_jong_generalized_2008}. We will not provide any further comment on the cost-estimation problem in the remainder of this paper; we take the solution to this problem as given. Instead, we focus on the \emph{bidding problem} - what offer ought we to make to a customer, given an estimate of the cost of insuring them? The optimal price to offer a customer depends not only on the estimated cost of insuring them, but additionally on, (1) the offers that other insurers on the PCW will make to the customer; (2) the customer's sensitivity to price differentials; and, (3) whether the customer is the type of client the insurance firm wishes to insure. Our work focuses on this third aspect the bidding problem. 

Throughout our paper, we use the term \emph{portfolio} to refer to the set of customers who accept an insurance firm's offers over some specified period of time (for example a day, week, or month). The problem of \emph{portfolio optimisation} is that of deciding what the ideal portfolio would look like; how many of each different type of customer the insurance firm wishes to have concurrently insured. This may depend on aspects such as the kind of brand and reputation the firm wishes to cultivate, the desire to maintain a diverse customer base so as to protect against correlated risk or systematic model inaccuracies, or the necessity to avoid insuring too many customers and being unable to settle all the incoming claims. We take the firm's solution to the problem of \emph{portfolio optimisation} also as given; that is, we suppose that the insurance firm already has a specification for the distribution of customers it would like to insure over some relevant future time-period. However, unlike in the stock market, where an investor can easily achieve their desired portfolio of stocks by simply investing and divesting, an insurance firm cannot arbitrarily add and remove customers from its portfolio. Rather, the firm must wait until a desired customer is asking for a quote, and then modulate the offer made to the customer in order to increase the likelihood that the offer is accepted. In deciding how to modulate the offer, the firm must carefully balance the value of having the customer in the portfolio against the profit they expect to make from insuring the customer. It would not be effective to, for example, make desired customers offers at which the firm is, on average, guaranteed a substantial loss. We term the problem of how to modulate offers made to customers in order to achieve a desired portfolio the problem of \emph{portfolio pursuit}. 

It is to the problem of \emph{portfolio pursuit} to which we make our contribution. Note that -- unlike the cost-estimation, bidding, and portfolio optimisation problems -- portfolio pursuit is intrinsically sequential in nature. If we wish to insure no more than 500 customers over the next week, the offers we ought to make to customers on Sunday depend heavily on our success rate for Monday through Saturday. As such, we cast the portfolio pursuit problem as a Markov Decision Process (MDP), and derive a novel Reinforcement Learning (RL) algorithm for its solution. Our decomposition into the four sub-problems given above - cost-estimation, pricing, portfolio optimisation, and portfolio pursuit - additionally enables us to develop a methodology which is minimally disruptive, in the sense that it can be integrated naturally with arbitrary solutions to the other three sub-problems. Indeed, our algorithm requires only a slight adjustment to existing methodologies for the bidding problem. 

\subsection*{Previous work}

To our knowledge, our paper is the first work to articulate a general mathematical formalism of the portfolio pursuit problem in insurance as a sequential decision making problem, and attempt to address it using reinforcement learning. \cite{treetanthiploet_insurance_2023} and \cite{krasheninnikova_reinforcement_2019} both apply reinforcement learning to other aspects of bid optimisation in the context of insurance. Like us, \cite{treetanthiploet_insurance_2023} considers the new business market for insurance. They have no notion of a desired portfolio, thus making each decision purely local. As such, they treat the problem as a contextual bandit, with no sequential aspect (beyond learning). In contrast, \cite{krasheninnikova_reinforcement_2019} examine the problem of pricing at renewal. While they consider the customer portfolio (thereby making the problem sequential), they restrict themselves only to constraints on the size of the portfolio, formulating it as a constrained Markov Decision Process. We have built upon their work by additionally factoring in the internal structure of customer portfolios, thereby giving a more general and flexible approach to portfolio pursuit. 

Many previous works have attempted to apply reinforcement learning to the portfolio optimisation problem in the context of other markets, such as the stock market \citep{hieu_deep_2020, benhamou_deep_2021, yang_deep_2023} and the cryptocurrency market \citep{jiang_deep_2017}.  However, as noted above, this situation is fundamentally different from the case considered here. In these other markets, the relevant financial products can typically be purchased very quickly, allowing one to shift funds between them more or less freely (subject to transaction costs). This is in contrast to customers, who arrive sequentially and over a prolonged period. Furthermore, while purchasing cryptocurrency or a stock is a fairly simple transaction, bidding for a customer is not, since it involves attempting to out-compete alternative offers while accounting for stochasticity in a customer's decision. As such, this previous literature is not directly applicable to the problem under consider here. 

\subsection*{Our contribution}

We propose a novel reinforcement learning algorithm for the portfolio pursuit problem over long time horizons. We cast this as an MDP, and give a novel algorithm for its solution. We take a model-based approach, allowing our algorithm to be trained entirely offline, thereby protecting against risk in deployment. The methodology we use can be integrated naturally into existing machine learning pipelines for insurance pricing. Unlike previous research addressing using RL for the purposes of insurance pricing \citep{krasheninnikova_reinforcement_2019, treetanthiploet_insurance_2023}, our insurance agent acts in a realistic agent-based environment \citep{macal_everything_2016, england_agent-based_2022}. This environment includes randomly generated customers along with a collection of other insurance agents that iteratively change their pricing strategies over time, creating a dynamic, non-stationary environment. The full code implementing both the market environment and our reinforcement learning solution is available in the \href{https://github.com/EdwardJamesYoung/RL-portfolio-pursuit}{code repository} accompanying this paper. 

The rest of this paper is organised as follows. In \secref{sec:problem formulation} we give a mathematical formulation of the \emph{portfolio pursuit problem}. In \secref{sec:methods} we describe our solution to the problem. We begin by describing a standard pipeline for solving the bidding problem, and the models used within that pipeline. We then outline a baseline methodology intended to mimic a naive approach to portfolio pursuit that may currently be used within industry. Lastly, we specify our approach, giving the full algorithm used for training, and compare the method to existing RL algorithms. In \secref{sec:results} we show the performance of our method against the baseline, demonstrating that our method is able to generate more profit than the baseline methodology while not compromising the quality of the final portfolio. Finally, in \secref{sec:discussion} we outline some limitations of our method and give opportunities for future research. 

\section{Problem formulation}\label{sec:problem formulation}

\subsection{Portfolio pursuit as a Markov Decision Process}

In our market environment, an epoch consists of $T$ interactions between a customer and the insurance agent. The customers $s_1,\dots,s_T$ themselves are drawn independently from the same underlying customer distribution. At each step, in response to the current customer features, the insurer takes an action $a_t \in [1,2]$. The action $a$ corresponds to an offer of $C(s) \times a$, where $C(s)$ is the expected cost of serving the customer\footnote{As noted in the introduction, we assume this as given. In practice, the expected cost of serving a customer is not known in advance, but must instead be estimated using historic claims data. However, this estimation problem is distinct from the problem of how to price insurance, given this estimate, which is what this paper focuses on.}. In response to the offer made by the insurer, and the offers made by other insurers within the market, the customer either accepts the offer or rejects it. If the offer is rejected ($y_t = 0$) then no profit is obtained by the insurer. If the offer is accepted $(y_t = 1)$ then an (expected) profit of $\Pr(s_t,a_t)$ is obtained. The profit is simply $\Pr(s,a) = C(s)a - C(s) = C(s)(a - 1)$. When the offer is accepted $(y_t = 1)$ then the customer is added to the insurer's portfolio\footnote{In reality, there may be a delay between an offer being made and accepted - indeed, customers often have up to 14 days to decide whether to accept an offer. In this case, a proxy measure, such as click-through from the PCW to the insurer's website, may be used in place.}. The portfolio at the $t$-th time-step is $\rho_t$, and the insurer's portfolio before the first customer $\rho_1$ is initialised as empty\footnote{The choice to initialise the portfolio as empty corresponds to optimising the portfolio of \emph{new} customers obtained over some time horizon. If instead the firm is interested in its \emph{total} portfolio - including all existing customers - the portfolio can be initialised to that instead, with only minor adjustments required elsewhere in our method.}, \ie $\rho_1 = \{\}$.

The performance of the insurer at the final time step is evaluated according to the net profit achieved minus a loss which measures the discrepancy between the final and desired portfolio\footnote{Note that this encompasses the case where the loss is a function of only the final portfolio.}, \ie
\begin{equation}
	\sum_{t=1}^T y_t \Pr(s_t,a_t) - \lambda \L(\rho_{T+1}, \rho^*).
\end{equation}
Here, $\lambda$ is a coefficient which controls the relative importance of profit vs. pursuit of a particular portfolio. 

In the MDP formalism, the states consist of customer, portfolio, and time-step triples, $(s_t, \rho_t, t)$. The reward signal following action $a_t$ is $y_t \Pr(s_t,a_t)$, with an additional reward $-\lambda \L(\rho_{T+1}, \rho^*)$ administered after the final customer interaction.

\subsection{Specification of portfolio loss function}\label{sec:portfolio loss function}

We stress that the algorithm we present in \secref{sec:RL for PP} makes no assumptions whatsoever about the nature of the loss function $\L$. However, for concreteness in our numerical experiments, we adopt one particular formulation of the loss. 

The loss function used in our numerical experiment is defined via a collection of $I$ indicators functions $\1_{A_i}, i = 1,\dots,I$ over the customer features. Each indicator is specified by set $A_i, i = 1,\dots,I$ in the space of customer features. These indicators allow us to represent portfolios as vectors in $\R^I$, with the $i$-th component given by $f_i = \sum_{s \in \rho} \1_{A_i}(s)$, where $\1_{A_i}(s)$ is equal to $1$ if $s \in A_i$, and $0$ otherwise. Given these vector representations of portfolios, we can implement the loss $\L$ via any distance measure on $\R^I$. Note that in practice we specify the target portfolio $\rho^*$ only through its vector representation $f^*_i \defeq \sum_{s \in \rho^*} \1_{A_i}(s)$; similarly, we will track the insurer's portfolio only via its frequency representation. For our numerical experiments we adopt the loss function:
\begin{equation}
	\mathcal{L}(\bm{f}, \bm{f}^*) \defeq \frac{1}{I} \sum_{i=1}^I  \frac{ |f_i - f^*_i| }{ \max(f_i, f_i^*) }.
\end{equation}
We choose this loss for the following properties: it is zero precisely when $f_i = f^*_i$; it is bounded between $[0,1]$; and it is symmetric in $\bm{f}$ and $\bm{f}^*$. Note that because the loss is bounded between $0$ and $1$, the coefficient $\lambda$ can be interpreted as answering the question: if the firm had the worst possible portfolio, how much short-term profit would they be willing to give up in order to trade it for the best possible portfolio? 

Our reinforcement learning algorithm (detailed in \secref{sec:RL for PP}) trivially extends to alternative loss functions which provide other ways of quantifying the distance between the obtained and target portfolio. This includes, for example, adding weights to each indicator to penalise discrepancies along different feature sets differently, or using an optimal transport distance to provide a more detailed measure of differences between sets. We leave exploration of alternative loss formulations to future work. 

\section{Methods}\label{sec:methods}

\subsection{Standard industry methodology}\label{sec:standard industry method}

We compare our method to a pipeline which mimics current industry approaches to insurance pricing and price modulation. In this section we describe a methodology which does not make reference to a target portfolio. This pipeline mimics standard industry pricing methods. We will then modify this method in two ways - firstly, to mimic standard industry methodology for modulating prices to attain a target portfolio, and secondly according to our new proposed algorithm.  

The insurance firm is assumed to have a dataset of interactions from previous epochs. Each entry in this interaction dataset includes:
\begin{enumerate}
	\item The customer features, $s_t$
	\item The expected cost of serving the customer, $C(s_t)$
	\item The action taken by the insurance firm, $a_t$
	\item The offers made by other insurance firms, $o_t$
	\item Whether the customer accepted the offer of the firm, $y_t$
\end{enumerate}
From the offers made by other firms, $o_t$, and the expected cost, $C(s_t)$, the insurer calculates \emph{market variables}, $m_t$. The market variables for the insurer are the average of the top-1, top-3, and top-5 offers made to the customer by other firms within the market, normalised by dividing by the expected cost $C(s_t)$ of serving the customer. Note that the market variables are not available to the insurer at inference time, but become available after the fact through the purchasing of historic data from PCWs. 

From this dataset, the insurer trains the following models:
\begin{enumerate}
	\item The \emph{market model} $M$, which maps from customer features, $s_t$, to market variables $m_t$, $s_t \mapsto M(s_t) \approx m_t$.
	\item The \emph{conversion model} $p$, which maps from the market variables, $m_t$, and action, $a_t$, to the probability the customer accepts the offer, $(m_t,a_t) \mapsto p(m_t,a_t) \approx \P(y_t = 1|m_t, a_t)$.
	\item The \emph{action model}, or \emph{policy} $\pi$, which maps from market variables, $m_t$, to \emph{optimal} actions $a^*_t$, $m_t \mapsto \pi(m_t) \approx a^*_t$.
\end{enumerate}
For our numerical simulations, we used a random forest for the market model, a Multi-Layer Perceptron (MLP) for the conversion model, and a Gaussian Process with Matern kernel for the action models \citep{rasmussen_gaussian_2005}. Our overall methodology places no restrictions on any of the models, although in practice we found it better to use continuous conversion and action models. If it is necessary to ensure these models are interpretable, simpler models - such as Generalised Linear Models (GLMs) - may be used instead of MLPs \citep{de_jong_generalized_2008}. More expressive models may work better for larger scale simulations, with correspondingly larger datasets, than those explored here. 

The market model $M$ and conversion model $p$ are trained directly on the historic dataset available to the insurer. To train the action model, we use \emph{regression-based amortised optimisation} \citep{amos_tutorial_2023}. Firstly, note that the expected profit, given that the customer accepts the insurer's offer, is $C(s_t)(a_t - 1)$. Accordingly, the insurer's expected profit, for action $a_t$, given market variables $m_t$, is $C(s_t)(a_t - 1)p(m_t, a_t)$. Therefore, for a random sample of market variables $m_t$ from the dataset, we use a one-dimensional optimiser to find $a^*_t \in \arg\max_a p(m_t, a)(a - 1)$ within the range $a_t \in [1, 2]$. We then fit the action model $\pi$ to $500$ $(m_t, a^*_t)$ pairs. 

At inference time, the features for a customer $s_t$ are fed into a market model to obtain estimates for the market variables\footnote{As the market variables are predictive of customer behaviour, and available historically but not at inference time, the use of a market model can be seen as a form of data imputation.}, $m_t = M(s_t)$. These are then fed into the action model to obtain the action $a_t = \pi(m_t)$. An offer of $C(s_t)a_t$ is then made to the customer. On $12\%$ of customers we explore by multiplying the offer by a randomly chosen modulation factor sampled from $[0.93, 0.95, 0.97, 1.03, 1.05, 1.07]$. 

Before continuing, we briefly remark that in practice the training and use of auxiliary models may be more complicated than presented here. To begin with, information about whether a customer accepts an offer is not available instantaneously, but only after a ``cooling-off'' period, typically 14 days. Likewise, there may be a delay in obtaining information about market variables for a given customer. Finally, insurers may choose to forgo the use of an action model all-together, instead solving the one-dimensional optimisation problem directly at inference time (rather than amortising as we do here). None of these observations present substantial difficulty to the RL algorithm we describe below. Indeed, our approach can be seen as a methodology which accepts as input auxiliary models and adjusts the prices returned by those models to solve the portfolio pursuit problem. It is therefore agnostic as to both the types of models used, and the data which those models are trained on. 

\subsection{Baseline methods for portfolio pursuit}\label{sec:baseline method}

It is standard practice to modulate the output prices offered to customers from the bidding model according to other features of interest for the customer, corresponding to the customer's desirability within the portfolio. This modulation is typically performed using domain experts' knowledge, rather than according to a data-driven or formal scheme. We attempt here to give a method which mimics this type of modulation as a baseline against which to compare our method, fully recognising that it does not accurately capture the complexity of standard industry practices. We encourage future work to implement more realistic and nuanced methodologies, and have facilitated this by making our codebase implementing the market environment publicly available. 

The baseline method makes use of historical assessments of the frequency with which the insurer encountered customers of the type relevant to their portfolio. In particular, for each indicator set $A_i$, compute the frequency representation $f_i = \sum_{s \in \rho_{T+1}} \1_{A_i}(s)$ for terminal portfolios $\rho_{T+1}$ for epochs during which there was no portfolio-based action modulation, \ie where actions were chosen via the method described in the previous section, \secref{sec:standard industry method}. We average this value over such historical epochs, giving us a frequency $\bar{f}_i$. $\bar{f}_i$ therefore acts as an estimate of how many customers within set $A_i$ we are likely to get \emph{without} action modulation. We compare this to the desired frequency, $f^*_i = \sum_{s \in \rho^*} \1_{A_i}(s)$. For each incoming customer, we look at each set $A_i$ that they call into, and compute the corresponding ratio of frequencies, $\bar{f}_i/f^*_i$. If this ratio is greater than 1 it indicates that, in order to achieve the desired portfolio, we should increase the offer price made to that customer. If the ratio is less than $1$, we should decrease the offer price. At inference time, we therefore modulate the offer price made to a customer $s$ by multiplying the action by
\begin{equation}
	\prod_{i: s \in A_i} g\left( \frac{\bar{f}_i}{f^*_i} \right).
\end{equation}
The map $g$ is given by 
\begin{equation}
	g(z) = 1 + \beta \frac{z^n - 1}{z^n + 1},
\end{equation} 
where $n, \beta \geq 0$. The functional form of $g$ was selected because it is monotonically increasing and bounded, with $g(z) \in [1 - \beta, 1 + \beta]$. The function parameters $n \in \{0.5, 1, 2\}$ and $\beta \in \{ 0.02, 0.05, 0.1, 0.25 \}$ are found by a grid-search; in our experiments we used values of $n = 1, \beta = 0.02$ for a loss coefficient of $\lambda = 2000$. Note that for a baseline method, $\beta, n$ must be chosen differently for each value of the loss coefficient $\lambda$. This is in contrast to our method, which applies equally well for any value of $\lambda$. 

\subsection{A Reinforcement Learning approach to Portfolio Pursuit}\label{sec:RL for PP}

\subsubsection*{Mathematical Preliminaries}

Our reinforcement learning algorithm uses a dynamic programming approach with function approximation to provide a policy which modulates customer prices according to the current portfolio possessed by the insurer. We define the \emph{portfolio value function}, $V_\pi(\rho, t)$, which gives the value of having portfolio $\rho$ at time $t$, given that future actions are selected by a (time-step dependent, deterministic) policy $\pi = \pi(s, \rho, t)$:
\begin{equation}\label{eq:portfolio value function}
	V_\pi(\rho, t) \defeq \E_\pi\left[ \sum_{k=t}^T y_k \Pr(s_k, a_k) - \lambda \L(\rho_{T+1}, \rho^*) | \rho_t = \rho \right].
\end{equation}
The corresponding \emph{action value function} $Q_\pi(s,a,\rho,t)$ is defined by:
\begin{equation}\label{eq:action value function}
	Q_\pi(s, a, \rho, t) \defeq \E_\pi\left[ \sum_{k=t}^T y_k \Pr(s_k, a_k) - \lambda \L(\rho_{T+1}, \rho^*) | \rho_t = \rho, s_t = s, a_t = a \right].
\end{equation}
The policy is optimal for all time steps if it satisfies the appropriate Bellman optimality equation, \ie
\begin{equation}\label{eq:bellman optimality}
	\pi(s, \rho, t) \in \arg\max_a Q_\pi(s, a, \rho, t).
\end{equation}
In \appref{app:bellman recursion}, we show by expanding the action value function that \eqref{eq:bellman optimality} is equivalent to:
\begin{equation}\label{eq:optimal action with values}
	\pi(s, \rho, t) \in \arg\max_a \P(y_t = 1 |s_t = s, a_t = a) \left( a - k_\pi(s,\rho, t+1) \right),
\end{equation} 
where $k_\pi(s, \rho, t+1)$ is defined by:
\begin{equation}\label{eq:k definition}
	k_\pi(s, \rho, t) \defeq 1 - \frac{V_\pi(\rho\cup \{s\}, t) - V_\pi(\rho, t)}{C(s)}.
\end{equation}
An important consequence of the above result is that the optimal policy is a function only of the customer features $s_t$ and the value $k_\pi$, and not of the portfolio or time-step directly except via $k_\pi$.

We also obtain (see \appref{app:bellman recursion}) the following recursion relationship for the portfolio value function:
\begin{equation}\label{eq:portfolio value recursion}
	V_\pi(\rho, t) = V_\pi(\rho, t+1) + \E\left[  C(s) \P(y_t = 1 |s_t = s, a_t = \pi(s,\rho, t)) \left( \pi(s,\rho, t) - k_\pi(s, \rho, t+1) \right)  \right],
\end{equation}
with boundary condition $V_\pi(\rho, T+1) = - \lambda \L(\rho, \rho^*)$. Here, the expectation is taken over the customer distribution. 

\subsubsection*{Outline of our method}

As with the industry baseline method, our method will make use of a market, conversion, and action model. We will use the conversion model $p(m, a) \approx \P(y_t = 1 | m_t = m, a_t = a)$ in place of the $\P(y_t = 1|s_t = s, a_t = a)$ in both the value recursion relationship \eqref{eq:portfolio value recursion} and the optimal action relationship \eqref{eq:optimal action with values}\footnote{We stress that this is only an approximation. In reality -- owing to how we constructed the market environment -- the probability of acceptance also depends on various customer features.}. Additionally, our action model (or policy) $\pi$ will now be a function of both the market variables $m$ and the k-values $K$, $\pi(m,k)$. 

Our method begins by training these models in much the same manner as the pipeline described in \secref{sec:standard industry method}, namely by fitting models to a previously obtained dataset. However, our method for generating the dataset for fitting the action model $\pi$ is slightly different. We first sample market variables $m_t$ from our dataset. For each market variable sample, we independently sample a $k$-value, $k_t$, from a Laplace distribution with location $1$ and scale $0.1$. We then compute 
\begin{equation}\label{eq:optimal action objective}
	a^*_t \in \arg\max_a p(m_t, a)(a - k_t),
\end{equation} where $p$ is our conversion model. We then fit the action model to approximate $\pi(m_t, k_t) \approx a^*_t$. We used a sample of $500$ $(m, k, a^*)$-triples to fit our conversion model. We additionally train a value function $V_\phi(\rho, t)$, with parameters $\phi$, as outlined in the next section. 

At inference time, for a customer $s_t$ with expected cost $C(s_t)$ and portfolio $\rho_t$ at time $t$, we use \eqref{eq:k definition} to compute the $k$-value $k_t$ via:
\begin{equation}\label{eq:k inference equation}
	k_t = 1 - \frac{V_\phi(\rho_t\cup\{s_t\},t+1) - V_\phi(\rho_t,t+1)}{C(s_t)}.
\end{equation}
We then feed the customer features $s_t$ into the market model to obtain estimates for the market variable $m_t = M(s_t)$. We then use the bidding model to obtain the action $a_t = \pi(m_t, k_t)$. Finally, we make an offer of $C(s_t)a_t$ to the customer. 

\subsubsection*{Value function training}

In order to train our inference-time value function $V_\phi(\rho,t)$, we make use of an \emph{next-step value function}, $U(\rho)$. This function is used during training, but not at inference time. Throughout training, it approximates the values of portfolios at the next time step. $U$ is used to create value estimates, which are then stored in a dataset. Our value function $V_\phi$ (which will be used at inference time) is then fit to this dataset. As discussed in \secref{sec:portfolio loss function}, portfolios are specified only through their frequency representation, $\rho \mapsto (f_i)_{i=1}^I, f_i = \sum_{s \in \rho} \1_{A_i}(s)$, since this preserves all information relevant for computing the loss. Accordingly, both $U$ and $V_\phi$ act on frequency representations, \ie vectors in $\R^I$. In the remainder of this section we do not distinguish between a portfolio and its frequency representation. 

We iterate backwards through time, setting $t = T,\dots,1$. At each time step, we use $U(\rho)$ to approximate $V_\pi(\rho, t+1)$. Initially, we set $U(\rho) = -\lambda \mathcal{L}(\rho, \rho^*) = V_\pi(\rho, T+1)$. At each time step, we use the next-step value function $U$ to form value estimates via \eqref{eq:k definition} and \eqref{eq:portfolio value recursion}. To approximate the expectation over the customer distribution in \eqref{eq:portfolio value recursion}, we compute an empirical average over samples from a \emph{customer reply buffer}, which stores the features, market variables, and expected cost of previous customers. We then fit $U$ to these value estimates thus obtained, such that $U$ now approximates $V_\pi(\rho, t)$. Having fit $U$, we sample an augmenting set of portfolios, and use $U$ to estimate each of their values. We finally store the value estimates in a dataset, $\mathcal{D}$, and then iterate backwards, $t \gets t - 1$. Once we have completed the backwards iteration through time, we fit our value function $V_\phi$ to the value estimates found in $\mathcal{D}$.

Before fitting, we apply a trick to make our dataset easier to learn. Notice that at inference time, $V_\phi$ is used only to compute $k$ values via \eqref{eq:k inference equation}. However, this equation is invariant to shifting all values at same time step $t$ by the same amount. Therefore, before fitting $V_\phi$, we recentre all the value estimates $V$ at each time step by subtracting the mean value estimate, $\bar{V}$.  We then fit our value function $V_\phi$ to the residuals, $\hat{V} = V - \bar{V}$. We found that this step is critical to effective learning, as it requires the value function $V_\phi$ to learn the \emph{differences} in value between portfolios, rather than overall magnitude. These differences are what drives decision making - indeed, the centring of our value estimates makes $V_\phi$ analogous to an advantage function used in other RL methods. 

Our approach is summarised in \algref{alg:training value function}. In our experiments, we set $U$ to be linear model, and $V_\phi$ to be an MLP. As with the conversion, market, and action models, nothing in our algorithm necessitates the use of these particular models for $U$ and $V_\phi$. We use $N = 500$ samples from the customer reply buffer, and $J = 24$ portfolio samples for fitting the next-step model, and an additional $J^+ = 120$ augmenting portfolios. 

\begin{algorithm}
	\caption{Training the value function $V$}\label{alg:training value function}
	\begin{algorithmic}[0]
		\State Initialise $U(\rho) \gets - \lambda \mathcal{L}(\rho, \rho^*)$. \Comment{Boundary condition for values.}
		\State Initialise empty dataset $\mathcal{D} \gets \emptyset$.
		\For{$t = T, \dots, 1$}
			\State Sample portfolios $\{ \rho_j \}_{j=1}^J$. 
			\For{$j=1,\dots,J$}
				\For{$i=1,\dots,N$}
					\State Sample $(s_i,m_i,C(s_i))$ from the customer reply buffer.
					\State Compute $k$ value: $k_i \gets 1 - \frac{U(\rho_j\cup\{s_i\}) - U(\rho_j)}{C(s_i)}$. \Comment{Using \eqref{eq:k definition}}
					\State Compute action: $a_i \gets \pi(m_i,k_i)$.
				\EndFor
				\State Compute value estimate for $\rho_j$: $V_j \gets U(\rho_j) + \frac{1}{N} \sum_{i=1}^N C(s_i) p(a_i, m_i)(a_i - k_i)$. \Comment{Using \eqref{eq:portfolio value recursion}.}
			\EndFor
			\State Fit $U$ to the input-output pairs $(\rho_j, V_j)$. \Comment{U now approximates $V_\pi(\rho, t)$.}
			\State Sample augmenting portfolio set $\{ \rho^+_j  \}_{j=1}^{J^+}$.
			\State Create $t$-dataset $\mathcal{D}_t \gets \{ (\rho_j, t, V_j) \}_{j=1}^J \cup \{ (\rho^+_j, t, U( \rho^+_j ) ) \}_{j=1}^{J^+}$.
			\State Recentre values in $\mathcal{D}_t$: $\hat{V} \gets V - \bar{V}$ for $\bar{V} = \frac{1}{|\mathcal{D}_t|} \sum_{(\rho,t,V) \in \mathcal{D}_t} V$.
			\State Add points to the main dataset: $\mathcal{D} \gets \mathcal{D}\cup \mathcal{D}_t$.
		\EndFor
		\State Fit value function $V_\phi$ to inputs $\rho,t$ and outputs $\hat{V}$ from $\mathcal{D}$.
	\end{algorithmic}
\end{algorithm}

In our algorithm, there are two points at which we sample portfolios. We have not yet specified how this sampling occurs. Ideally, these portfolios would be \emph{on-policy}, \ie the portfolios sampled for time step $t$ would be taken from the distribution of portfolios our policy encounters at time $t$. However, since we do not have access to any such data, our method is forced to be off-policy. The sampling of portfolios comes from three sources: \emph{previously on-policy portfolios}, \emph{target on-policy portfolios}, and \emph{high-coverage portfolios}. The sampling is done in a ratio of 1:1:2. We explain each of these methods in turn. 

Much like the baseline method discussed in \secref{sec:baseline method}, the \emph{previously on-policy portfolio sampling} makes use of historic data about the frequency with which customers in each category appear naturally \ie without action modulation. As before, we denote by $\bar{f}_i$ the average of the frequency representations $\sum_{s \in \rho_{t+1}} \1_{A_i}(s)$ over historic epochs in which no portfolio was being pursued. We can then define a historic rate, given by $\bar{p}_i = \bar{f}_i/T$. To sample previously on-policy portfolios at time $t$, we sample each component of the portfolio's frequency representation independently from the Binomial distribution with parameters $p_i + 1/T, t$. The addition of $1/T$ to the rate was made to ensure that, for each category, we have a strictly positive probability of sampling portfolios with customers in that category. Sampling these portfolios ensures we have a good representation of value for portfolios that we would encounter if we did not adjust our pricing methodology. The \emph{target on-policy portfolio sampling} method works in the same manner as the previously on-policy portfolio sampling, except we use $p^*_i = f^*_i/T$ where $f^*_i = \sum_{s \in \rho^*} \1_{A_i}(s)$ is the target frequency. 

The last method we use is \emph{high-coverage portfolios}. This method is parametrised by a \emph{slope parameter}, $\sigma \in [0,1]$. We first define $\bm{p}_{\max} = (1 + \sigma)\max( \bm{p}^*, \bar{\bm{p}} ), \bm{p}_{\min} = (1 - \sigma)\min( \bm{p}^*, \bar{\bm{p}} )$, where the maximum and minimum are taken element-wise. At time $t$, we consider evenly spaced points $\omega \in [0,1]$. For each $\omega$ point, we construct the frequency representation $\bm{f}_t(\omega) = t( \omega \bm{p}_{\max} + (1 - \omega)\bm{p}_{\min} )$, which interpolates between the frequency representation if customers were acquired at either the maximum rate or minimum rate. This will in general not be integer valued, so we therefore take the integer part of each element, and then add on a Bernoulli sample which is one with probability equal to the remainder term. The high-coverage portfolios thus sampled ensures a wide sampling outside the on-policy distributions. The contribution of these portfolios is crucial, as it ensures our method still has accurate assessments of portfolio values outside the on-policy distribution, providing stability at inference time. We used a value of $\sigma = 0.9$.

\subsection{Relationship to existing RL methods}

Here we briefly discuss the relationship of our proposed RL algorithm to existing algorithms within the literature. Our method can be viewed as an \emph{actor-critic} algorithm, in the sense that it learns both a policy $\pi(m,k)$ and a value function $V_\phi(\rho,t)$. However, our method displays substantial differences to more conventional actor-critic methods, such as A3C \citep{mnih_asynchronous_2016}, PPO \citep{schulman_proximal_2017}, SAC \citep{haarnoja_soft_2018}, and DDPG \citep{lillicrap_continuous_2019}. 

Firstly, our method is \emph{not} a policy-gradient algorithm. Unlike the methods listed above, our policy is not learned by finding the gradient of a value measure (such as total return \citep{mnih_asynchronous_2016}, or a $Q$-function \citep{lillicrap_continuous_2019}) with respect to the policy parameters. Instead, we exploit the fact that - because the action space is one-dimensional - the relevant value measure, given in \eqref{eq:optimal action objective}, can be optimised directly with a non-gradient-based optimiser. We then fit our policy directly to the optimised values - a form of regression-based amortised optimisation \citep{amos_tutorial_2023}. This allows us to avoid problems relating to local maxima which arise in gradient-based methods. It also frees us from the necessity of having a policy which is differentiable in the parameters; indeed, in our experiments we use a Gaussian process as our policy \citep{rasmussen_gaussian_2005}. In fitting the policy to a dataset of actions which maximise a value measure, our method is akin to the cross-entropy method \citep{simmons-edler_q-learning_2019}. 

Secondly, our value function is not a function of the full MDP state. The full MDP state is the (customer, portfolio, time-step) triple, $(s,\rho,t)$. Instead, our value function only depends on the portfolio and the time-step, $(\rho,t)$. This substantially reduces the dimensionality of the input space to our value function, since the number of customer features is (typically) much larger than the number of portfolio categories. From \eqref{eq:optimal action with values}, we know that the optimal action depends only on the next portfolio, time-step pair. This means that, for the purposes of action-selection, it suffices to learn only these values, rather than the full MDP state values (or, indeed, the value of state-action pairs \citep{mnih_human-level_2015, haarnoja_soft_2018, lillicrap_continuous_2019}). We also note that the dependence is only through the $k$-value. This enables us to learn a policy which takes $k$ directly as input, rather than the full MDP state. Importantly, this policy can be learned fully before learning any value functions (as we do in our algorithm), and does not require re-training while the value function is learned, unlike a typical actor-critic method \citep{mnih_asynchronous_2016,lillicrap_continuous_2019}. 

Lastly, our method utilises a memory-reply buffer to act as a model of the arrival of new customers. However, unlike many other methods which use a memory reply buffer for value-learning \citep{mnih_human-level_2015, haarnoja_soft_2018, lillicrap_continuous_2019}, we do not sample states randomly. Rather, (portfolio, time-step) pairs, $(\rho, t)$, are sampled systematically by iterating $t$ backwards in time. In iterating backwards, our method is analogous to a dynamic programming algorithm, which starts at terminal states of known value (in our case, terminal portfolios) and applies backwards induction to find the value of all states recursively. Similar to the Fitted Q-Iteration algorithm \citep{ernst_tree-based_2005}, our method avoids the curse of dimensionality associated with exact dynamic programming in MDPs by approximating the value function at each time-step by a linear model, $U$. Because we are applying an approximate dynamic programming method we need only sweep through the state-space once. This is in contrast to methods which apply a Bellman optimality equation but do not recurse backwards from terminal states \citep{mnih_human-level_2015}, where each state must be sampled many times in order to approximate the correct value. 

\iffalse 

\subsubsection{Other applications of our algorithm}\label{sec:other applications}

Although we have designed our algorithm specifically for the portfolio pursuit application pursued in this paper, certain elements readily generalise to other situations not considered here. We briefly enumerate potential other uses of elements of our algorithm. 

Firstly, our algorithm depends on the fact our state-space factorises into portfolio, time-step, customer feature triples $(\rho, t, s)$. The fact that we have a fixed horizon $T$, and that the time-step $t$ is a component of the state-space, allows us to compute values by iterating backwards in time. 

the use of a simple (in our case, linear) model 

\fi

\subsection{Additional simulation details}\label{sec:additional simulation details}

Our simulations use a realistic agent-based model of an insurance market, mediate through a PCW \citep{macal_everything_2016, england_agent-based_2022}. The insurance firm implementing each method -- baseline (\secref{sec:baseline method}) or RL (\secref{sec:RL for PP}) -- competes with $5$ other insurance firms (none of which are pursuing a portfolio, as in \secref{sec:standard industry method}). Each customer is randomly generated by sampling features from a complex generative distribution, with realistic interdependencies between features. For example, customer income depends on the location, age, and occupation of the customer. Only a subset of features are revealed to the insurers though the PCW interface. On the basis of these features, each insurer makes the customer an offer. The customer then either selects an offer or chooses to walk away from the market. Full details of the market environment can be found in the \href{https://github.com/EdwardJamesYoung/RL-portfolio-pursuit}{code repository} that accompanies this paper. 

Target portfolios are generated at random by sampling $I = 5$ customer features, and setting the target quantity $f_i^*$ for those features to be twice $\bar{f}_i$ if $\bar{f}_i \leq 10$, and otherwise half or twice of $\bar{f}_i$ at random (rounding up or down respectively to ensure integer values). We have $6$ burn-in epochs, during which historic data is gathered for training models. Insurers use the past 4 epochs to train models, and all burn-in epochs for computing historical frequency representations. For the first burn-in epoch, before any models are trained, insurers take actions randomly and uniformly in $[1,1.2]$. Following the burn-in epochs, we then have $8$ testing epochs. Each epoch consists of $T = 1000$ steps. No re-training of any models occurs following a testing epoch, meaning the testing epochs are independent trials of the same value function $V_\phi$. In the results below, we aggregate statistics across testing epochs. We set the loss-coefficient $\lambda = 2000$. The results shown in \secref{sec:results} are obtained by aggregating statistics across both the $8$ testing epochs and $24$ independent trials, with each trial having a different randomly generated target portfolios. To ensure a fair comparison between the RL and baseline methods, we use the same set of random seeds for both methods. This means that, up until the testing epochs, both methods interact with the market identically, ensuring the same historic data to train on and target portfolio to aim towards. Thus, any differences in performance are attributable only to differences over the testing epoch. 

\section{Results}\label{sec:results}

\begin{figure}[!h]
	\includegraphics[width=0.9\textwidth]{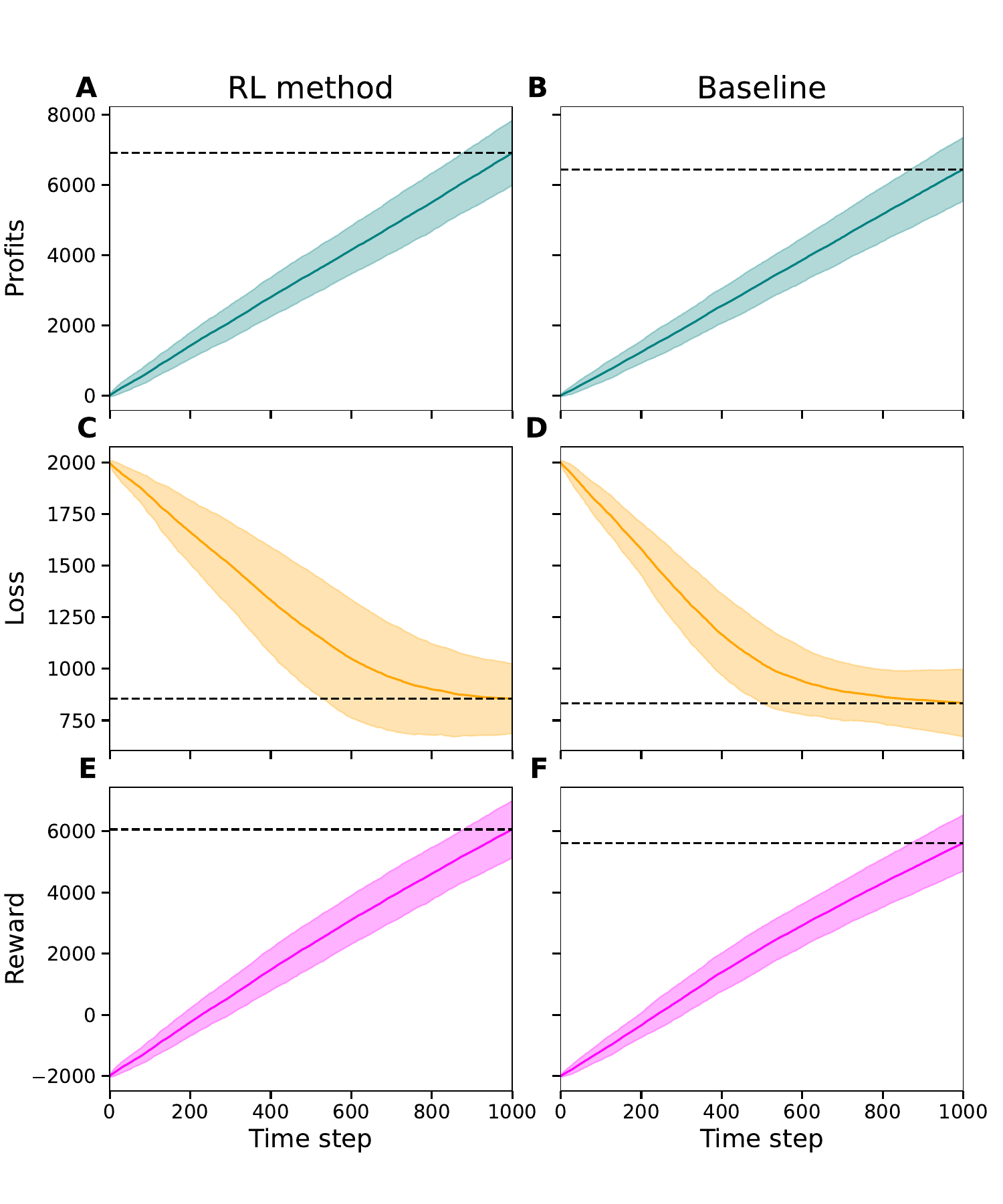}
	\caption{\textbf{A side-by-side comparison of our method with an industry baseline}. We plot the mean (with standard deviation error bars) of three quantities of interest over the testing epoch: (A-B) the profit $\sum_{k=1}^t y_k \Pr(s_k,a_k)$, (C-D) the loss between the current and target portfolio $\lambda \L(\rho_t, \rho^*)$, and (E-F) the difference between them. Left-hand panels (A,C,E) correspond to our method, \secref{sec:RL for PP}, and right-hand panels (B, D, F) correspond to the baseline method, \secref{sec:baseline method}. The black dashed line in each panel corresponds to the mean value at the terminal time-step. See \secref{sec:additional simulation details} for further information about our experimental set-up.}
	\label{fig:main-results}
\end{figure}

\figref{fig:main-results} shows the results of comparing our method to the baseline averaged over 24 trials. On each trial, both methods have the same interaction history, and thus have the same training data, up until the testing epochs. They also pursue identical target portfolios. Thus, differences in performance stem only from different behaviour on the test epochs. From panels (A-B) we see that, on average, our method generates more profit than the baseline (one-sided Mann-Whitney U test, $p < 0.0001$; $\mathrm{CLES} = 65\%$; $\mathrm{Cohen's~d} = 0.51$, moderate effect size). Over the course of $1000$ customer interactions, on average our method generates an additional $\pounds 472$ (or a $7\%$ increase) in profit. Panels (C-D) show the loss between the target portfolio and the obtained portfolio at each time step. The final loss distribution of both methods is very similar (two-sided Mann-Whitney U test, $p = 0.16$, not significant), showing that our method has not compromised portfolio quality in order to generate additional profit. Thus we can conclude that our method brings in additional revenue while getting comparably close to the desired portfolio as the baseline. Panels (E-F) show the total reward (profit minus loss) of both methods, which is the objective that both methods are optimising. We see that our method outperforms the baseline, achieving higher reward on average over the trials (one-sided Mann-Whitney U test, $p < 0.0001$; $\mathrm{CLES} = 63\%$; $\mathrm{Cohen's~d} = 0.48$, moderate effect size). In Table \ref{table:table of results} we tabulate the average final profit, loss, and reward of both methods.  

\begin{table}[h!]
	\centering
	\begin{tabular}{|c ||c c | c|} 
		\hline
		 & Profit & Loss & Reward \\ 
		\hline\hline
		Reinforcement learning method & 6916 & 856 & 6060 \\
		Baseline method & 6444 & 834 & 5610 \\ \hline
		RL minus baseline & 472 & 22 & 450 \\  
		\hline
	\end{tabular}
	\caption{\textbf{The performance of both the reinforcement learning and baseline methods on key metrics}. Across the columns we show the final profit, $\sum_{k=1}^T y_k \Pr(s_k,a_k)$, loss $\lambda \mathcal{L}(\rho_{T+1},\rho^*)$, and total reward, $\sum_{k=1}^T y_k \Pr(s_k,a_k) - \lambda \mathcal{L}(\rho_{T+1},\rho^*)$. Across rows we show the metric values for the reinforcement learning agent, baseline agent, and the difference between them.}
	\label{table:table of results}
\end{table}

\section{Discussion}\label{sec:discussion}

\subsection*{Limitations and opportunities for future work}

We believe that our work presents an important first step in the application of ideas from reinforcement learning to the portfolio pursuit problem. However, our work has several limitations, with corresponding opportunities for future research. 

In our investigation, we limited the time span over which insurers pursued a particular portfolio to only $T = 1000$ time-steps, with one customer arriving per time-step. This is a relatively short length of time for a large insurer, who may see orders of magnitude more customers per day. Our choice to limit the number of time-steps was motivated primarily by the prohibitive computational cost of running dozens of market simulations for larger numbers of time steps. Preliminary experiments provided evidence that our method will likely continue to work for larger numbers of time-steps. However, for much larger lengths of time, \eg on the order of months, a different approach may well be needed. Accordingly, we believe that the framework and methods we have develop above could be integrated naturally with RL methods which attempt to perform temporal abstraction, \eg goal-conditioned policies \citep{nasiriany_planning_2019} or the options framework \citep{precup_temporal_2000}. Indeed, longer time-scale portfolio pursuit can be accomplished by chaining together shorter time-scale portfolio pursuit with the appropriately chosen intermediate target portfolios. We leave the problem of selecting such intermediate target portfolios to future research. 

In our framework, we have also made the simplifying assumption that customers do not leave the portfolio once they have entered. This is justified when the number of time-steps is small, since few customers will both accept and then cancel a contract over the course of, \eg one day. However, for longer time-scales, it may begin to become necessary to model the exit of customers from the portfolio. In \appref{app:customers leaving} we expand our analytic derivation to consider this case, and provide some preliminary suggestions for how to integrate this into our algorithm. 

Lastly, our method is model-based, with optimal actions and values computed using a conversion model which estimates the probability of customers accepting any given offer. Consequently, if the conversion model is incorrect, our bidding strategy will accordingly be sub-optimal. For example, the bidding model may underestimate the probability that a customer accepts an offer at a given price, and make offers too low. Model uncertainty could be incorporated into the algorithm by explicitly penalising the value of actions about which the insurer is uncertain. A core step in making the algorithm more robust would be allowing for online updates of the models in light of new data coming in. This is an increasingly pressing problem within insurance pricing in particular, since such markets tend to be very non-stationary, shifting rapidly as individual insurers change their pricing strategy and inflation drives changes in prices. An effective pricing strategy will need to precisely trade-off between profit and the value of the information generated by making particular offers. Although we do not address it in this paper, we expect this to be a fruitful area for future research into the application of RL to insurance pricing. 

\subsection*{Conclusion}

Our paper has introduced a new formalism for a problem within the insurance industry - that of portfolio pursuit. Portfolio pursuit asks the question of how to modulate the offers made to individual consumers, in light of a desired target portfolio which describes the higher-level strategy of the insurance firm. We formulated portfolio pursuit it as an MDP in \secref{sec:problem formulation}. Given its formal description, in \secref{sec:methods} we discussed both industry standard methodologies for addressing this problem, and gave our own reinforcement learning solution. We then showed in \secref{sec:results} that our method outperforms the industry baseline, generating more profit while getting equally close to the target portfolio. It is our hope that this paper will stimulate additional research into this area, laying the groundwork for better algorithms and methodologies for portfolio pursuit. 

\subsection*{Acknowledgements}

EJY is supported by the UKRI Engineering and Physical Sciences Research Council Doctoral Training Program grant EP/T517847/1. Work was completed during an internship at Accenture (UK) Limited, in connection with the Turing Institute. 

\subsection*{Declaration of competing interest}

The authors declare no competing interests.

\subsection*{Codebase}

The GitHub code repository associated with this paper can be found \href{https://github.com/EdwardJamesYoung/RL-portfolio-pursuit}{here}. The code is licensed under Creative Commons Attribution-NonCommercial 4.0 International Public License (CC BY-NC 4.0)

\subsection*{Author contributions}

Problem statement was formulated jointly by EJY, AR, and JJ. Market environment designed by EJY. Mathematical derivations, algorithm design, and numerical experiments were preformed by EJY. EJY drafted the manuscript, with additional comments and feedback provided by ET, AR, and JJ. 

\newpage

\bibliographystyle{apalike}
\bibliography{RL-portfolio-citations}

\begin{thebibliography}{}

\bibitem[Amos, 2023]{amos_tutorial_2023}
Amos, B. (2023).
\newblock Tutorial on {Amortized} {Optimization}.
\newblock {\em Found. Trends Mach. Learn.}, 16(5):592--732.

\bibitem[Benhamou et~al., 2021]{benhamou_deep_2021}
Benhamou, E., Saltiel, D., Ohana, J.~J., Atif, J., and Laraki, R. (2021).
\newblock Deep {Reinforcement} {Learning} ({DRL}) for {Portfolio} {Allocation}.
\newblock In Dong, Y., Ifrim, G., Mladenić, D., Saunders, C., and Van~Hoecke,
  S., editors, {\em Machine {Learning} and {Knowledge} {Discovery} in
  {Databases}. {Applied} {Data} {Science} and {Demo} {Track}}, pages 527--531,
  Cham. Springer International Publishing.

\bibitem[de~Jong and Heller, 2008]{de_jong_generalized_2008}
de~Jong, P. and Heller, G.~Z. (2008).
\newblock {\em Generalized {Linear} {Models} for {Insurance} {Data}}.
\newblock International {Series} on {Actuarial} {Science}. Cambridge University
  Press, Cambridge.

\bibitem[England et~al., 2022]{england_agent-based_2022}
England, R., Owadally, I., and Wright, D. (2022).
\newblock An {Agent}-{Based} {Model} of {Motor} {Insurance} {Customer}
  {Behaviour} in the {UK} with {Word} of {Mouth}.
\newblock {\em Journal of Artificial Societies and Social Simulation}, 25(2).
\newblock Number: 2 Publisher: University of Surrey.

\bibitem[Ernst et~al., 2005]{ernst_tree-based_2005}
Ernst, D., Geurts, P., and Wehenkel, L. (2005).
\newblock Tree-{Based} {Batch} {Mode} {Reinforcement} {Learning}.
\newblock {\em J. Mach. Learn. Res.}, 6:503--556.

\bibitem[Haarnoja et~al., 2018]{haarnoja_soft_2018}
Haarnoja, T., Zhou, A., Abbeel, P., and Levine, S. (2018).
\newblock Soft {Actor}-{Critic}: {Off}-{Policy} {Maximum} {Entropy} {Deep}
  {Reinforcement} {Learning} with a {Stochastic} {Actor}.
\newblock arXiv:1801.01290 [cs, stat].

\bibitem[Hieu, 2020]{hieu_deep_2020}
Hieu, L.~T. (2020).
\newblock Deep {Reinforcement} {Learning} for {Stock} {Portfolio}
  {Optimization}.
\newblock {\em International Journal of Modeling and Optimization},
  10(5):139--144.
\newblock arXiv:2012.06325 [cs, math, q-fin].

\bibitem[Jiang et~al., 2017]{jiang_deep_2017}
Jiang, Z., Xu, D., and Liang, J. (2017).
\newblock A {Deep} {Reinforcement} {Learning} {Framework} for the {Financial}
  {Portfolio} {Management} {Problem}.
\newblock arXiv:1706.10059 [cs, q-fin].

\bibitem[Krasheninnikova et~al., 2019]{krasheninnikova_reinforcement_2019}
Krasheninnikova, E., García, J., Maestre, R., and Fernández, F. (2019).
\newblock Reinforcement learning for pricing strategy optimization in the
  insurance industry.
\newblock {\em Engineering Applications of Artificial Intelligence}, 80:8--19.

\bibitem[Lillicrap et~al., 2019]{lillicrap_continuous_2019}
Lillicrap, T.~P., Hunt, J.~J., Pritzel, A., Heess, N., Erez, T., Tassa, Y.,
  Silver, D., and Wierstra, D. (2019).
\newblock Continuous control with deep reinforcement learning.
\newblock arXiv:1509.02971 [cs, stat].

\bibitem[Macal, 2016]{macal_everything_2016}
Macal, C.~M. (2016).
\newblock Everything you need to know about agent-based modelling and
  simulation.
\newblock {\em Journal of Simulation}, 10(2):144--156.
\newblock Publisher: Taylor \& Francis \_eprint:
  https://doi.org/10.1057/jos.2016.7.

\bibitem[Mnih et~al., 2016]{mnih_asynchronous_2016}
Mnih, V., Badia, A.~P., Mirza, M., Graves, A., Lillicrap, T.~P., Harley, T.,
  Silver, D., and Kavukcuoglu, K. (2016).
\newblock Asynchronous {Methods} for {Deep} {Reinforcement} {Learning}.
\newblock arXiv:1602.01783 [cs].

\bibitem[Mnih et~al., 2015]{mnih_human-level_2015}
Mnih, V., Kavukcuoglu, K., Silver, D., Rusu, A.~A., Veness, J., Bellemare,
  M.~G., Graves, A., Riedmiller, M., Fidjeland, A.~K., Ostrovski, G., Petersen,
  S., Beattie, C., Sadik, A., Antonoglou, I., King, H., Kumaran, D., Wierstra,
  D., Legg, S., and Hassabis, D. (2015).
\newblock Human-level control through deep reinforcement learning.
\newblock {\em Nature}, 518(7540):529--533.
\newblock Publisher: Nature Publishing Group.

\bibitem[Nasiriany et~al., 2019]{nasiriany_planning_2019}
Nasiriany, S., Pong, V.~H., Lin, S., and Levine, S. (2019).
\newblock Planning with {Goal}-{Conditioned} {Policies}.
\newblock arXiv:1911.08453 [cs, stat].

\bibitem[Precup, 2000]{precup_temporal_2000}
Precup, D. (2000).
\newblock {\em Temporal abstraction in reinforcement learning}.
\newblock phd, University of Massachusetts Amherst.
\newblock AAI9978540 ISBN-10: 0599844884.

\bibitem[Rasmussen and Williams, 2005]{rasmussen_gaussian_2005}
Rasmussen, C.~E. and Williams, C. K.~I. (2005).
\newblock {\em Gaussian {Processes} for {Machine} {Learning}}.
\newblock The MIT Press.

\bibitem[Schulman et~al., 2017]{schulman_proximal_2017}
Schulman, J., Wolski, F., Dhariwal, P., Radford, A., and Klimov, O. (2017).
\newblock Proximal {Policy} {Optimization} {Algorithms}.
\newblock arXiv:1707.06347 [cs].

\bibitem[Simmons-Edler et~al., 2019]{simmons-edler_q-learning_2019}
Simmons-Edler, R., Eisner, B., Mitchell, E., Seung, S., and Lee, D. (2019).
\newblock Q-{Learning} for {Continuous} {Actions} with {Cross}-{Entropy}
  {Guided} {Policies}.
\newblock arXiv:1903.10605 [cs].

\bibitem[Treetanthiploet et~al., 2023]{treetanthiploet_insurance_2023}
Treetanthiploet, T., Zhang, Y., Szpruch, L., Bowers-Barnard, I., Ridley, H.,
  Hickey, J., and Pearce, C. (2023).
\newblock Insurance pricing on price comparison websites via reinforcement
  learning.
\newblock arXiv:2308.06935 [cs, q-fin].

\bibitem[Yang, 2023]{yang_deep_2023}
Yang, S. (2023).
\newblock Deep reinforcement learning for portfolio management.
\newblock {\em Knowledge-Based Systems}, 278:110905.

\end{thebibliography}

\newpage

\appendix

\section{Bellman recursions for portfolio values}\label{app:bellman recursion}

In this appendix we expand out the Bellman equations for the portfolio and action value functions, given by \eqref{eq:portfolio value function} and \eqref{eq:action value function} respectively. We start with the action value function. We will expand this in terms of the portfolio value function. Firstly, we denote the returns by
\begin{equation}
	G_t = \sum_{k=t}^T y_k \Pr(s_k, a_k) - \lambda \L(\rho_{T+1}, \rho^*).
\end{equation}
Then we have
\begin{align*}
	Q_\pi(s, a, \rho, t) &\defeq \E_\pi\left[ G_t | \rho_t = \rho, s_t = s, a_t = a \right] \\
	&= \P(y_t = 1 |s_t = s, a_t = a)\E\left[ G_t | \rho_t = \rho, s_t = s, a_t = a, y_t = 1 \right] \\ 
	&+ (1 - \P(y_t = 1 |s_t = s, a_t = a))\E\left[ G_t | \rho_t = \rho, s_{t+1} = s, a_{t+1} = a, y_t = 0 \right] \\
	&= \P(y_t = 1 |s_t = s, a_t = a)\left( \Pr(s,a) + \E\left[ G_{t+1} | \rho_{t+1} = \rho \cup \{ s \}  \right]  \right) \\
	&+ (1 - \P(y_t = 1 |s_t = s, a_t = a))\E\left[ G_{t+1} | \rho_{t+1} = \rho \right] \\
	&=\P(y_t = 1 |s_t = s, a_t = a)\left( \Pr(s,a) + V_\pi \left( \rho \cup \{ s \}, t+1 \right) \right) \\
	&+ (1 -\P(y_t = 1 |s_t = s, a_t = a)) V_\pi\left( \rho, t+1 \right) \\ 
	&= V_\pi(\rho, t+1) + \P(y_t = 1 |s_t = s, a_t = a) \left( \Pr(s,a) + V_\pi \left( \rho \cup \{ s \}, t+1 \right) -  V_\pi\left( \rho, t+1 \right) \right).
\end{align*}
Note that, since the profit is the offer price $C(s)a$ minus the expected cost $C(s)$, the profit is thus given by: $\Pr(s,a) = C(s)(a - 1)$. We can then say that:
\begin{align*}
	Q_\pi(s,a,\rho,t) &= V_\pi(\rho, t+1) + \P(y_t = 1 |s_t = s, a_t = a) \left( C(s)(a - 1) + V_\pi \left( \rho \cup \{ s \}, t+1 \right) -  V_\pi\left( \rho, t+1 \right) \right) \\
	&= V_\pi(\rho, t+1) + C(s) \P(y_t = 1 |s_t = s, a_t = a) \left( a - k_\pi(s,\rho, t+1) \right), 
\end{align*}
where $k_\pi(s, \rho, t+1)$ is defined in \eqref{eq:k definition}. Since the first term is a constant in $a$, and $C(s) > 0$, this shows that maximisation of $Q_\pi$ over $a$ is equivalent to maximising:
\begin{equation}
	\P(y_t = 1 |s_t = s, a_t = a) \left( a - k_\pi(s,\rho, t+1) \right).
\end{equation} 
Furthermore, note that $V_\pi(\rho, t) = \E\left[ Q_\pi(s,a,\rho, t) \right]$. Assuming a deterministic policy $\pi = \pi(s,\rho, t)$, this gives us the following recursion:
\begin{equation}
	V_\pi(\rho, t) = V_\pi(\rho, t+1) + \E\left[  C(s) \P(y_t = 1 |s_t = s, a_t = \pi(s,\rho, t)) \left( \pi(s,\rho, t) - k_\pi(s, \rho, t+1) \right)  \right]. 
\end{equation}

\section{Generalisation to customers leaving}\label{app:customers leaving}

In this section we generalise the analysis is \appref{app:bellman recursion} to the case where customers can leave the portfolio. 

We define the action-value function as:
\begin{equation}
	Q_\pi(s, a, \rho, t) \defeq \E_\pi\left[ G_t | \rho_t = \rho, s_t = s, a_t = a \right] 
\end{equation}
We will now compute the Bellman recursion for $Q_\pi$. To do so, we introduce a portfolio transition distribution, $p( \rho' | \rho )$. This gives the probability of having portfolio $\rho'$ at time $t+1$, given that you \emph{ended} time-step $t$ with portfolio $\rho$. 
\begin{align*}
	Q_\pi(s, a, \rho, t) &= \P(y_t = 1 |s_t = s, a_t = a)\E\left[ G_t | \rho_t = \rho, s_t = s, a_t = a, y_t = 1 \right] \\ 
	&+ (1 - \P(y_t = 1 |s_t = s, a_t = a))\E\left[ G_t | \rho_t = \rho, s_{t+1} = s, a_{t+1} = a, y_t = 0 \right] \\
	&= \P(y_t = 1 |s_t = s, a_t = a)\left( \Pr(s,a) + \int \E\left[ G_{t+1} | \rho_{t+1} = \rho' \right] p(\rho' | \rho \cup \{s\} ) d\rho' \right) \\
	&+ (1 - \P(y_t = 1 |s_t = s, a_t = a)) \int \E\left[ G_{t+1} | \rho_{t+1} = \rho' \right] p(\rho' | \rho  ) d\rho' \\
	&=\P(y_t = 1 |s_t = s, a_t = a)\left( \Pr(s,a) +\int V_\pi \left( \rho', t+1 \right) p(\rho' | \rho \cup \{s\}) d\rho' \right) \\
	&+ (1 -\P(y_t = 1 |s_t = s, a_t = a)) \int V_\pi\left( \rho', t+1 \right) p(\rho' | \rho  ) d\rho' \\ 
	&= \int V_\pi\left( \rho', t+1 \right) p(\rho' | \rho  ) d\rho' \\ 
	&+ \P(y_t = 1 |s_t = s, a_t = a) \left( \Pr(s,a) + \int V_\pi \left( \rho', t+1 \right) \left[ p(\rho'|\rho \cup \{s\}) - p(\rho'| \rho) \right] d\rho' \right) \\
	&= \E\left[ V_\pi(\rho', t+1) | \rho \right] \\
	&+ \P(y_t = 1| s_t = s, a_t = a) \left( \Pr(s,a) + \E\left[ V_\pi(\rho', t+1) | \rho\cup \{ s \} \right] - \E\left[ V_\pi(\rho', t+1) | \rho \right] \right)
\end{align*}
We now use the substitution that $\Pr(s,a) = C(s)(a - 1)$. Following the derivation in \appref{app:bellman recursion}, we define: 
\begin{equation}\label{eq:k definition with leaving}
	k_\pi(s, \rho, t) = 1 - \frac{\E\left[ V_\pi(\rho', t) | \rho\cup \{ s \} \right] - \E\left[ V_\pi(\rho', t) | \rho \right]}{C(s)}
\end{equation}
Our optimal policy is then the same. Integrating over customers and actions, we obtain the following recursion for the portfolio value function $V_\pi$: 
\begin{equation}\label{eq:portfolio value recursion with leaving}
	V_\pi(\rho, t) = \E\left[ V_\pi(\rho', t+1) | \rho \right] + \E\left[ C(s) \P(y_t=1|s_t=1,a_t = \pi(s,\rho,t))(\pi(s,\rho,t) - k_\pi(s, \rho, t+1)) \right]
\end{equation} 
Given access to $p(\rho'|\rho)$, we can easily adapt \algref{alg:training value function} to use this recursion instead, by sampling portfolios from $p$ and averaging to approximate the expectations in \eqref{eq:k definition with leaving} and \eqref{eq:portfolio value recursion with leaving}. This adds some computational overhead, but is not prohibitive. The transition model $p(\rho'|\rho)$ can be learned from historic data. 

One issue we encounter when we account for the possibility of customers leaving the portfolio is that the portfolio objects become considerably more complicated. Recall from \secref{sec:RL for PP} that, when customers do not leave the portfolio, we can specify portfolios only through their frequency representation, since (i) it was unimportant when a customer joined the portfolio, and (ii) the other customer features were unimportant. However, in this case, neither of this conditions are true. Firstly, the probability of any given customer leaving depends how long they have been in the portfolio. Secondly, the probability of a customer leaving after being in the portfolio for any length of time may be a function of their other features. This necessitates a more detailed and possibly high-dimensional portfolio representation. We do not explore these issues further in our paper. 

\end{document}